# A Novel Feature Descriptor for Image Retrieval by Combining Modified Color Histogram and Diagonally Symmetric Co-occurrence Texture Pattern

[a]Ayan Kumar Bhunia, [b]Avirup Bhattacharyya, [c]Prithaj Banerjee, [d]Partha Pratim Roy*, [e]Subrahmanyam Murala

[a]Dept. of ECE, Institute of Engineering & Management, Kolkata, India Email- [a]ayanbhunia007@gmail.com
[b]Dept. of ECE, Institute of Engineering & Management, Kolkata, India. Email- [b] avirupiem@gmail.com
[c]Dept. of CSE, Institute of Engineering & Management, Kolkata, India. Email-[c]prithajtutanbanerjee@gmail.com
[d]Dept. of CSE, Indian Institute of Technology Roorkee, India. Email- [d]proy.fcs@iitr.ac.in
[e]Dept. of EE, Indian Institute of Technology Ropar, India. Email- [e]subbumurala@iitrpr.ac.in
[d]*email: proy.fcs@iitr.ac.in, TEL: +91-1332-284816*

## Abstract

In this paper, we have proposed a novel feature descriptors combining color and texture information collectively. In our proposed color descriptor component, the inter-channel relationship between Hue (H) and Saturation (S) channels in the HSV color space has been explored which was not done earlier. We have quantized the H channel into a number of bins and performed the voting with saturation values and vice versa by following a principle similar to that of the HOG descriptor, where orientation of the gradient is quantized into a certain number of bins and voting is done with gradient magnitude. This helps us to study the nature of variation of saturation with variation in Hue and nature of variation of Hue with the variation in saturation. The texture component of our descriptor considers the co-occurrence relationship between the pixels symmetric about both the diagonals of a 3×3 window. Our work is inspired from the work done by Dubey et al.[1]. These two components, viz. color and texture information individually perform better than existing texture and color descriptors. Moreover, when concatenated the proposed descriptors provide significant improvement over existing descriptors for content base color image retrieval. The proposed descriptor has been tested for image retrieval on five databases, including texture image databases - MIT VisTex database and Salzburg texture database and natural scene databases Corel 1K, Corel 5K and Corel 10K. The precision and recall values experimented on these databases are compared with some state-of-art local patterns. The proposed method provided satisfactory results from the experiments.

# 1. Introduction

Modernization in technology has led to advancement in various areas of academics, medicine, forensic analysis, entertainment, and other such developments. In the areas of image processing, this has led to the rising necessity of information retrieval from images. Significant amount of research has been done in this area using various methods to retrieve text and related information from images. Image recognition has found wide applications in various real time applications, for examples tracking automobiles on CCTV cameras, or helping blind people to travel, etc. For this work the main hindrance that is experienced, is from quality of images. Blurry images or images with a lot of contour deformations provide challenging scenarios for text detection. Lighting conditions, complex backgrounds are other such challenges that need to be overcome for proper text detection in images. So, to drive out these difficulties in image retrieval based on text information, content based image retrieval (CBIR) was introduced. In content-based image retrieval (CBIR) textual descriptions are avoided for image retrieval. Instead, similarities in their contents (textures, colors, shapes, etc.) with respect to the query image are considered and retrieval of images are done by listing images from large databases in descending order of similarity. There have been numerous researches done in content based image retrieval[2]–[10] in recent past involving both color and texture features.

Color quantization is closely related with color models. A large number of color models have been proposed and used for image retrieval and other related tasks over the years. Selecting the appropriate color space thus plays an important role in this aspect. The color model which is most commonly used in image processing and computer vision problems is the RGB model which contains three color channels, namely (R channel), Green (G channel) and Blue (B channel). Apart from texture, color information also serves to play an important role in content based image retrieval task [11]–[14] since it provides the global information about the image in terms of distribution of various color components. The drawback of using RGB color model is that the color information contained in the three channels is highly correlated. This prompted us to use the HSV color space in content based image retrieval framework in order to capture the color information efficiently. In the HSV color space, H, S and V stand for Hue, saturation and value, respectively. The Hue component is defined as an angle and it varies from 0 through 1 and the corresponding color varies from yellow, green, cyan, blue, magenta and back to red. Thus there are red values both at 0 and 1.0. The saturation is an indication of the purity of the color. Value component indicates the brightness which is almost similar to the gray scale version of RGB image.

In this paper, the primary motive of developing a feature descriptor is to efficiently capture both the color and texture information present in the image. This makes the feature a multipurpose descriptor that can work for a large variety of images belonging to different databases. Our work stresses on this point and aims at giving equally effective results on different image datasets available online. In this paper, we aim to develop a novel color descriptor by exploring the inter channel or the mutual relationship between H and S channels which has not been done in any previous work. Along with this, a texture descriptor has been designed which considers the relationship between the pixels symmetric about the left and right diagonals of a 3×3 window. These two descriptors individually perform better than existing color and texture descriptors and concatenation of these two gives significant improvement over existing descriptors for content base color image retrieval.

**1.1. Related Work**

Image retrieval process mainly focuses upon texture and color analysis. Local intensity of the image defines texture to some extent, which is why local neighborhood features and statistical features are discovered for such texture patterns and similarly color correlogram, color histogram, color coherence vector, etc. are used for low level color feature descriptor. Most renowned method for statistical feature extraction of images, Gray Level Co-occurrence Matrix (GLCM) was first proposed by Haralick [2]. The GLCM, also known as the Gray Level Spatial Dependence Matrix examines the texture by considering the spatial relationship of pixels. The texture of an image is characterized by the GLCM function by calculating co-occurrence of pairs of pixel with specific values and in a particular spatial relationship. A GLCM is thus created, and then from this matrix statistical measures are extracted. Features were calculated directly by applying GLCM to the texture image, then edge image was used by Zhang et al. for gathering even more concrete and relevant information [3]. Thus, GLCM of edge images was calculated and Prewitt edge detector was applied there in four directions. For LUV and RGB color channels, GLCM was further extended to single-channel co-occurrence matrix and then to multi-channel co-occurrence matrix. Application of color texture image retrieval [11] was thus introduced using GLCM. Gray Level Co-occurrence Matrix was used for retrieval of rock texture images by Partio et al. [4]. Siqueira et al. [5] utilized pyramid representation and Gaussian smoothing for multi-scale image extraction and retrieval. Some other applications of GLCM are [15]–[17].

For color and texture features proposition of integrated color and intensity co-occurrence matrix was done where composition of texture and color features were computed. Color representation was performed using HSV color space instead of RGB, and image retrieval was done by labelled and unlabelled image datasets [18]. In color histogram the frequency of every intensity is considered discarding the color spatial co-relation. This spatial co-relation is utilized in color correlogram [19]. Again, this color correlogram combined with supervised learning was used for feature vector extraction and thus improved result in image retrieval in two different ways, firstly by modifying the query image, and secondly by the distance metric learning [20]. For retrieving images, color coherence vector was proposed using image pixel color coherence and incoherence, and then it was compared with color histogram [13]. Technique of artificial neural network (ANN) was applied for image retrieval at faster rate by clustering images by Park et al. [21]. Quantization of color histogram for retrieving images was utilized in Gaussian Mixture Vector Quantization (GMVQ) [12]. The Motif Co-occurrence Matrix builds a 3D matrix, corresponding to local statistics of images was proposed for image retrieval [22]. Further an extension of this Motif Co-occurrence Matrix was used in Modified Color Motif Co-occurrence Matrix (MCMCM) for image retrieval using relationships between the color channels by Murala et al. [14]. Again, using HSV color space, text was used with motif matrix on histogram in [23].

Wavelet transform has found extensive application in description of image texture. Along the most prominent perceptual dimensions [24] the texture quality is determined. Wavelet transform was used [25] to collect texture and color features from an image. Daubechies' Wavelet Transform(DWT) was used for image searching and indexing in Wang et al. [26]. Here, the feature vectors are constructed by using the wavelet coefficients in the lowest few frequency bands and their variances. The idea of wavelet correlogram in Content Based Image Retrieval was first proposed in [27]. Information is extracted from an image only in three directions (horizontal, vertical and diagonal) by DWT. This directional limitation was removed by using Gabor wavelet feature based texture analysis in [28]. Gabor wavelet transform was used for texture classification in [29] by Ahmadian et al. Gabor wavelet correlogram is an extension of [27], which was proposed as a rotation invariant feature using Gabor wavelet in Content Based Image Retrieval [30].

Ojala et al. [31] first proposed the local binary pattern (LBP) for texture feature extraction. In LBP, an 8 bit binary string is used for representing the spatial relationship between the local neighboring pixels with its center pixel. Uniform version of LBP and rotation invariant LBP has been introduced for image classification and retrieval. Various extensions of LBP, e.g., Completed LBP (CLBP) [32], Block-based Local Binary Pattern (BLK LBP) [33], Dominant LBP (DLBP)[34], Center Symmetric LBP (CS-LBP) [35] etc., were introduced for image retrieval and texture classification. One major drawback of traditional LBP method is that the anisotropic features are not described in its circular sampling region. A Multi-structure local binary pattern (Ms-LBP) [36] operator was proposed as a solution to this problem, where an extended LBP operator was obtained by changing the shape of sampling region for texture image classification. In fact, for texture classification Gaussian as well as wavelet based low pass filters, were used in LBP called as Pyramid Local Binary Pattern(PLBP) [37] proposed by Qian et al. where multi-resolution images were extracted by using a low pass filter from the original image, and these multi-resolution low pass images are used in LBP features collection. Again, a combination of LBP along with the Gabor filter gave better result [38]. Besides these, the moment was applied in feature extraction in [39]. Again, the feature was extracted by the edge information in Directional Local Extrema Pattern [40]. There were various improvements made over LBP in Dominant Local Binary Pattern(DLBP) [34], Local Bit-plane Decoded Pattern(LBDP) [41], Local Edge Pattern for Segmentation and Image Retrieval (LEPSEG and LEPINV) [42], Local Mesh Pattern (LMP) [43], Average Local Binary Pattern(ALBP) [44] etc. To minimize noise effect in LBP numerous algorithms have been formulated. In Local Ternary Pattern (LTP) [45], firstly a threshold value is considered (say t) if the neighboring pixel values ($I_i$) is in the range of center pixel ($I_c$) $\pm$ threshold i.e. ($I_c - t, I_c + t$) then 0 is assigned and if it is less than this range -1 is assigned otherwise +1 is assigned. Then this ternary pattern is converted into upper and lower binary bit patterns and improved versions of LTP known as Improved LTP [46] gives better result. Noise-Resistant LBP (NR-LBP) [47], Robust LBP(RLBP) [48] are used in noise reduction of LBP feature. Second order derivation in horizontal direction as well as vertical directions are considered in Local Tetra Patterns [49] and it gives better result than LBP which is then transformed to binary patterns for calculations. Extended version of Local Tetra Patterns is Local Oppugnant Pattern [50] in RGB color space. Murala et al. proposed spherical symmetric 3D Local Ternary Patterns[51] using Gaussian filters and RGB color space which provided a 3D space and extracted LTP from every directions .

It is to be noted that our work is inspired by the work in [52] which used both color histogram and texture descriptor in order to capture the global and local information of the image respectively. However, to our knowledge none of the earlier local descriptors considered the inter channel relationship for histogram calculation in HSV color space. In other words, the mutual relationship between the channels has not been thoroughly investigated to evaluate the feature descriptor in content based image retrieval task. Also, the idea of exploring the relationship between diagonally symmetric pairs in a 3×3 window has not been done earlier. Following this, we develop one color histogram descriptor and one texture descriptor which upon concatenation provide significant improvement over the method in [52] and other existing methods as well.

The rest of the paper is organized as follows. In Section 1.2, the main contributions of our work are mentioned with respect to existing techniques. In Section 3, we detail the proposed color and texture descriptor. Section 4 presents the experimental results and advantages of the proposed descriptor. Finally, the conclusion part is mentioned in the last Section of the paper.

**1.2. Main Contributions**

A number of works [52]–[54] have focused on color content based image retrieval. The work in [53] proposed a novel image feature representation method using color information from the $L^*$ $a^*$ b* color space. They called it Color Difference Histogram (CDH) for image retrieval. Walia et al. [54] exploited the Color Difference Histogram (CDH) and Angular Radial Transform (ART) features to obtain color, texture and shape information of an image. They used a modified Color Difference Histogram to improve the retrieval performance. In [52], the authors simply quantized the histograms from the H and S channels into different bins and then concatenated those histograms to obtain the color feature. Very few works like the one by Lu et al. [55] developed a novel LBP-based color feature named Ternary-Color LBP (TCLBP), to represent the inter-channel information using the RGB color space. However, to the best of our knowledge, none of the existing works have exploited the inter-channel information existing in the HSV color space. In this paper, we exploit the inter-channel relationship between the H and S channels for color histogram computation. This is done by quantizing the H-channel (angle i.e. different color range) into different bins and voting using saturation value of the corresponding pixel position. This explores the inter channel or the mutual relationship between H and S channels in a novel manner. This is due to the fact that this novel histogram computation takes into account the actual saturation(S channel) values corresponding to a range of colors (a particular bin in the H channel) rather than just the number of occurrences of those values within the range. With the similar motivation, we quantize the S-channel into different bins and use the corresponding H-channel (different color range) value for voting.

A number of texture features based on local patterns have been proposed for texture based image retrieval by considering the relation among symmetric neighbors with respect to the center. The most popular among them is the Center Symmetric Local Binary Pattern (CSLBP) [35]. In CSLBP, the relation between the center symmetric pixels is considered for calculating the local pattern of the input image and the remaining neighbors are ignored. A modified form of CSLBP has been proposed by Verma et al. [52] for calculating the feature map. Moreover, they have calculated the texture feature using the V channel. Not many works have focused on exploring the mutual relationship between the diagonally symmetric neighbors though. The importance of representing the relationship between the diagonal neighbors was proposed in the work done by Dubey et al.[1]. Here, the first-order local diagonal derivatives are calculated for exploiting the relationship among the diagonal neighbors of a given center pixel in an image. The authors compared the intensity values of the center pixel with the intensity value of local diagonal for utilizing the relationship of the central pixel with its neighbors. Motivated from this work, we intend to explore the relationship among the diagonally symmetric neighbors along the left and right diagonals of a 3×3 window of an image. Moreover, we calculate the GLCM of the feature map obtained rather than computing the histogram to maintain the information of the spatial correlation.

The major contributions of our paper are as follows: Firstly, we introduce a novel method for color histogram calculation from H and S channels of HSV color space with an objective to explore the mutual relationship between the two channels. Secondly, a new texture descriptor is developed using the relationship between the diagonally symmetric neighbors along both the diagonals in a 3×3 window of an image. These two feature descriptors, color histogram and texture feature, are concatenated in order to utilize both the global and local information of the image respectively which found to be beneficial in our experiments. Thirdly, the resultant feature descriptor has been used for color image retrieval on different databases (Corel-1K, Corel-5K, Corel-10K database, Salzburg texture database and MIT VisTex database) and it has been found to be performing better than the method in [52] significantly and other existing methods as well.

## 2. Color and Texture Descriptor

### 2.1 Color histogram using inter-channel voting

Since the primary objective in this work is to exploit the Inter-channel relationship, we do not focus on separately quantizing the H and S channels into bins and concatenating histograms as done in [52]. Our principle is motivated from the popular HOG descriptor. The HOG (Histogram of Oriented Gradients) is a feature descriptor which calculates the gradient of an image in two different directions, X and Y. The orientation and magnitude of the gradient are calculated. The gradient vector is quantized into a histogram of P bins. Each individual bin is used to specify a particular octant in the angular space. The histogram is formed by adding the gradient magnitude g(x,y) to the bin indicated by quantized gradient orientation Ω(x,y). Similarly, our focus in this work is to quantize the Hue (holding color information) value $\emptyset(x, y)$ into different bins and adding up the Saturation value $S(x, y)$ to the bin indicated by $\emptyset(x, y)$. Studying the variation of the Hue with the Saturation is equally as important as studying the variation of Saturation with Hue. For the accomplishment of this objective, we consider it reasonable to quantize the Saturation value $S(x, y)$ into different bins and form the histogram by voting using the Hue values $\emptyset(x, y)$. If the Hue value at a particular pixel position (i,j) of the image be $\emptyset(i, j)$ and if it belongs to the $k^{th}$ quantized histogram bin, then we may write:

$$\text{Bin}(k) = \text{Bin}(k) + S(i, j) \ldots \ldots \ldots (1)$$

where, $S(i, j)$ is the saturation value at pixel position (i,j). Following the same principle, if we quantize the Saturation values into L bins and vote with the corresponding Hue value, we may write:

$$\text{Bin}(l) = \text{Bin}(l) + H(i, j) \ldots \ldots \ldots (2)$$

Thus, we construct two sets of histograms one with K bins and another with L bins to exploit the inter-channel relationship. The traditional histogram quantization method and our proposed histogram quantization method are shown in Fig. 1(a) and Fig. 1(b) respectively.

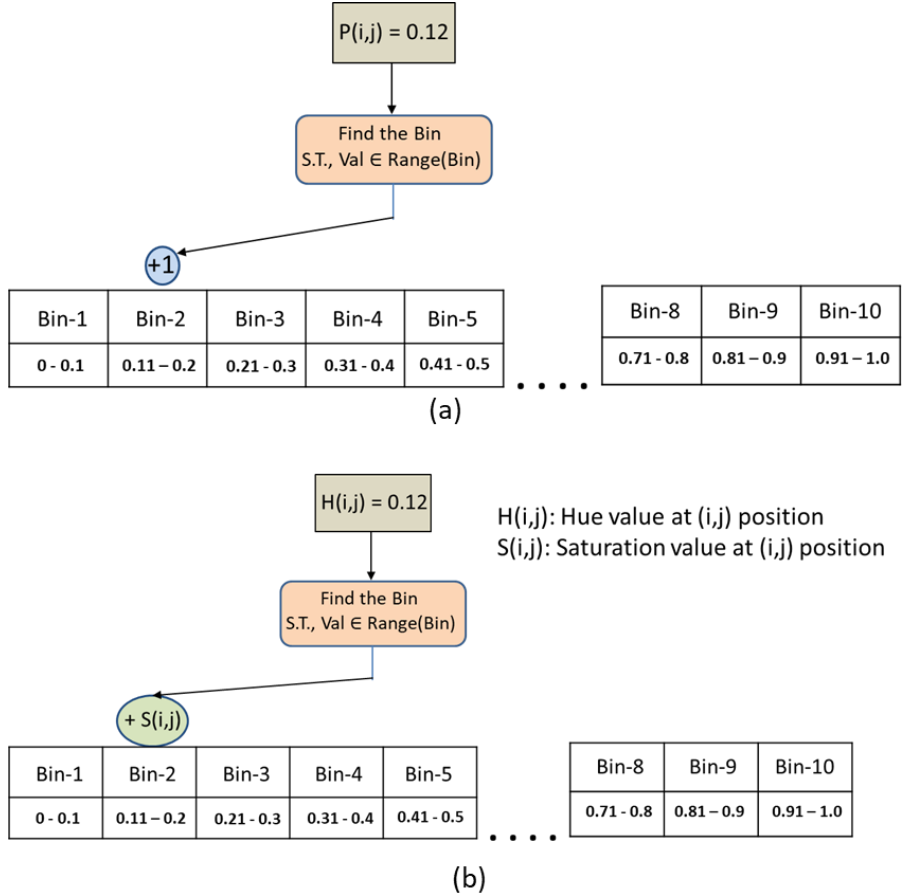

**Fig. 1. (a) Traditional histogram quantization method. (b) Our proposed histogram quantization method in order to explore the inter channel relationship.**

## 2.2 Local Pattern

### 2.2.1 Local Binary Patterns:

Ojala et al. proposed Local Binary Pattern (LBP) which was used mainly in texture classification [56] but its computational ease lead it to be further used in medical imaging [57], image classification [31], object tracking[58] and facial expression recognition [59]. This method uses a small window of an image. Here, each of the N neighboring pixels surrounding the center pixel is compared to the center pixel and a binary value (0 or 1) is assigned based on this intensity difference (as given in eqn. 3). The final result is obtained after multiplying these bits with specific weights. The center pixel is replaced with this value which is the binary pattern value for that center pixel. Thus by replacing each center pixel with its binary pattern value a local binary map of the image is generated in its gray level. A histogram of this local binary map is calculated to create the feature vector. Eqn. (3)-(6) gives the formula for LBP and the histogram.

$$LBP(N) = \sum_{k=1}^{N} 2^k \times \nabla_1(I_k, I_c) \quad \ldots \ldots \ldots (3)$$

$$\nabla_1(I_k, I_c) = \begin{cases} 1, & I_i \geq I_c \\ 0, & otherwise \end{cases} \ldots \ldots \ldots (4)$$

$$Hist(N)|_{LBP} = \sum_{k=1}^{X} \sum_{l=1}^{Y} \nabla_2(LBP(k,l), L); \quad L \in [0, (2^N - 1)] \quad \ldots \ldots \ldots (5)$$

$$\nabla_2(b_1, b_2) = \begin{cases} 1, & b_1 = b_2 \\ 0, & otherwise \end{cases} \ldots \ldots \ldots (6)$$

Here, N represent the number of neighboring pixels. The $k^{th}$ surrounding pixel is denoted by $I_k$ and center pixel is denoted by $I_c$. The final histogram of the pattern map is computed by eqn. (5). An example window for LBP calculation is given in Fig. 3(a).

### 2.2.2 Gray Level Co-occurrence Matrix

The concept of Gray Level Co-occurrence Matrix (GLCM) was proposed by Haralick et al. [2] in which they studied 24 features. It is used to study the co-occurrence of pixel pairs within a specific distance and in a particular direction in an image. It is a very popular statistical method for feature calculation. In this paper, we have calculated the GLCM of the feature map obtained after applying the proposed texture DSCoP (Diagonally Symmetric Co-occurrence Pattern) rather than computing a histogram. This has been done to exploit the spatial correlation of pixels in the feature map which is lost on histogram computation as it is purely a frequency distribution. The equation used for calculating GLCM of an input image:

$$M_d(l, m) = \#\{(a, b), (c, d): I(a, b) = l, I(c, d) = m\} \ldots \ldots \ldots (7)$$
$$\text{where } (a, b), (c, d) \in H_a \times H_b$$
$$(c, d) = a + k \times \emptyset_1, b + k \times \emptyset_2.$$

In this equation, $M_d$ is the gray level co-occurrence matrix. Here k represents the distance and $\emptyset$ represents the direction. $H_a \times H_b$ represents the horizontal and vertical spatial domains. $I(a, b)$ and $I(c, d)$ are the values of pixel intensity at positions $(a, b)$ and $(c, d)$. An example of GLCM calculation has been shown in Fig. 2. Fig. 2(a) shows the original matrix and in Fig. 2(b) we have calculated the GLCM of the matrix given in Fig. 2(a) with adjacent pairs (one-distance) and horizontal pixel pair (zero- degree direction).

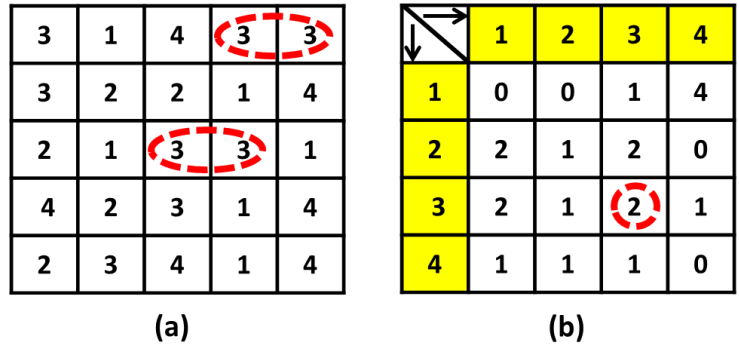

**Fig. 2**. Example showing the gray level co-occurrence pattern matrix calculation in (b) for matrix (a)

## 2.2.2 Diagonally symmetric co-occurrence pattern

In the present work, we have calculated a texture feature named Diagonally Symmetric Co-occurrence Pattern (DSCoP) for image retrieval. In this pattern, we consider the relationship between the diagonally symmetric neighboring pairs of a 3×3 window as shown in Fig. 3(b). There are two diagonals for every 3×3 window of an image. One is the principal diagonal and the second diagonal is the counter diagonal. If the symmetric neighbor pair about a given diagonal be $(I_k, I_j)$, where k={1,2,3,4,……,8} is one of the eight neighbors of the center and the center pixel be denoted by $I_c$ then $I'_k$ may be written as:

$$I'_k = I_k - I_c \qquad k = 1,2,...,8 \quad \ldots\ldots\ldots (8)$$

The values of k can be divided into two subsets. One set of values (k=1,7,8) for the principal diagonal and the second set of values (k=1,2,3) for the counter diagonal. As a result, the values of $I'_j$ may be expressed as:

$$I'_j = \begin{cases} I'_{\mathrm{mod}(12-k,8)}, & k = (1,7,8) \text{(for principal diagonal)} \\ I'_{8-k}, & k = (1,2,3) \text{(for counter diagonal)} \end{cases} \quad \ldots\ldots\ldots (9)$$

The relationship between $I'_k$ and $I'_j$ may be represented as:

$$\rho(I'_k, I'_j) = \begin{cases} 1, & I'_k \times I'_j \geq 0 \\ 0, & \text{else} \end{cases} \quad \ldots\ldots\ldots (10)$$

Thus, when both $I'_k$ and $I'_j$ are of the same sign, the resultant bit will be 1, otherwise it is zero. There are six neighbor pairs in total, three for each diagonal. We obtain a six bit binary string and calculate the decimal equivalent which replaces the center pixel of the window. Thereafter, we calculate the GLCM which helps to exploit the spatial co-relation. For GLCM computation we follow the same set of specifications as that followed in [52]. However, since we compute 64 feature vectors instead of 16, we quantize the GLCM into 16 levels from 0 to 15 to maintain the same feature dimension as used in [52].

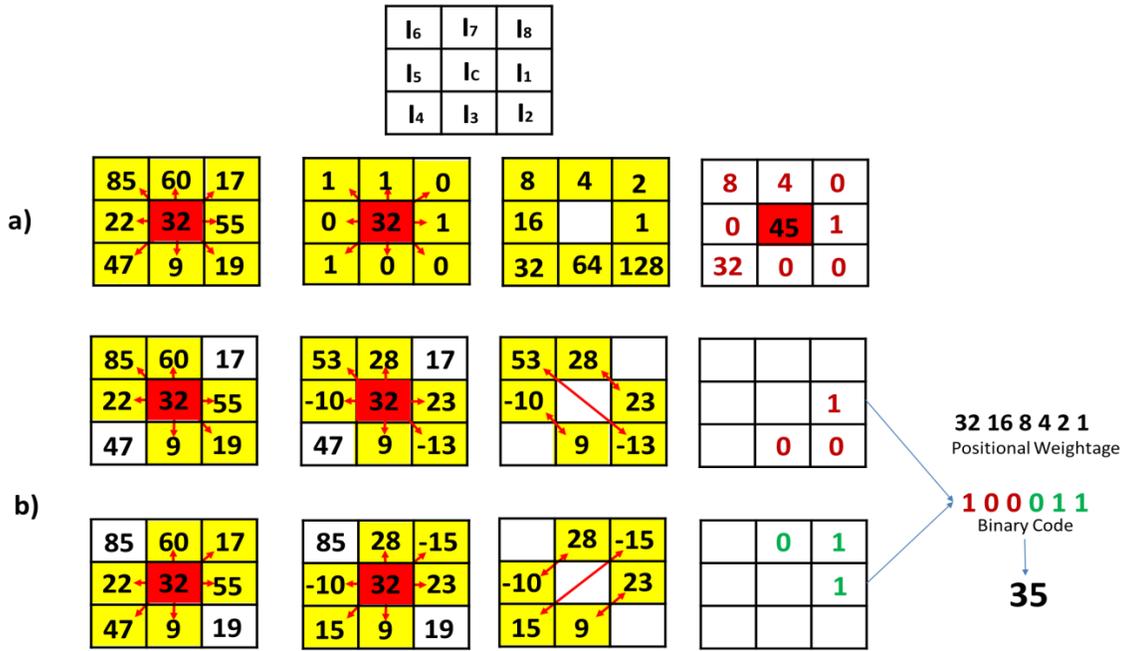

**Fig. 3. a) Example showing Local binary pattern calculation b) Illustration for Diagonally symmetric co-occurrence pattern calculation.**

### 2.3 Advantage of proposed descriptor:

1. The texture descriptor proposed in our work takes into account the relationship between the diagonally symmetric neighboring pairs about the principal and counter diagonal of a 3×3 window of an image rather than considering the relationship only between center symmetric pixels which has not been studied earlier in the literature so far.

2. A novel color descriptor by taking into consideration the inter channel relationship between the H and S channels of an image. This type of relationship between H and S channels has not been studied in the literature so far.

3. The proposed descriptor has been evaluated on a number of publicly available color and texture datasets. For each of them, the proposed descriptor has outperformed the existing descriptors for image retrieval.

## 3. Proposed System Framework

The proposed method has been illustrated with the help of a block diagram shown in Fig. 4, and the corresponding algorithm for the same in section 3.1.3. In this work, we have computed the color feature by quantizing the H channel into different bins and studied the variation of Saturation with respect to the Hue by following a principle similar to that of the HOG descriptor. Hue represents the color component and has a value between 0 and 1. For our experiments, the H channel has been divided into 18/36/72 bins. The variation of Hue with Saturation has also been studied following the same principle as mentioned above. For this purpose, the S channel has been quantized into 10/20 bins. All possible combinations of Hue and Saturation have been used and the results have reported in the results section. We have used the same value of normalization factor as used in [52] for all databases to appropriately justify the superiority of our method. For texture feature extraction, we have calculated the GLCM of the DSCoP pattern. The same set of specifications for GLCM computation as mentioned in [52] has been followed. The only difference is that we have quantized our GLCM matrix into 16 levels from 0-15 to obtain a 256 dimensional feature vector and keep the feature dimension same as [52].

The algorithm is shown in two parts. The first part describes the system framework. Here, an image is fed as input and in the output, the feature vector is obtained by concatenating the histograms generated by calculating the GLCM vector of DSCoP feature and the Modified Color Histogram. In part 2, retrieval of image is performed using the proposed feature extraction method. Here, query image is taken as input and in output the retrieved images are obtained based on similarity measure of the feature vectors as in part 1.

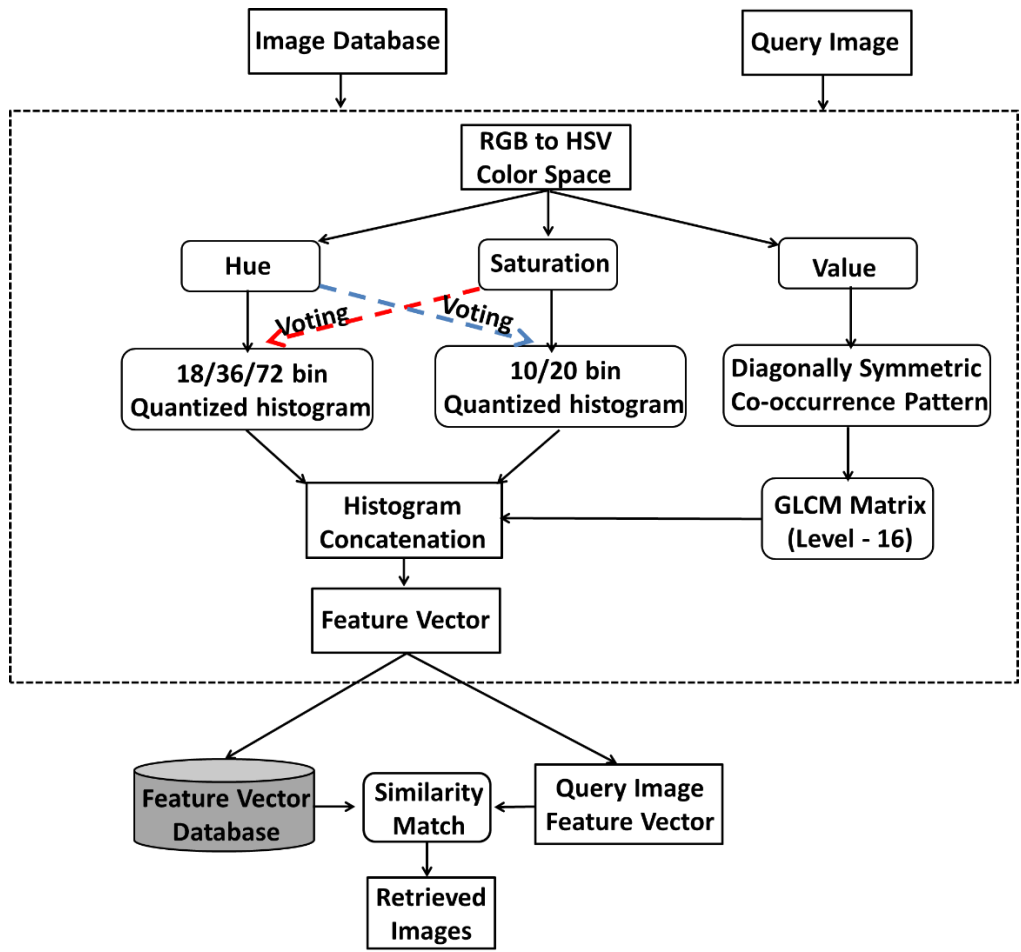

Fig. 4. Proposed system block diagram.

### 3.1. System Framework Algorithm:

**Part 1: Construction of feature vector**

**Input: An Image from the database**

**Output: Feature vector.**

1) Choose an image from database and convert it from RGB to HSV color space.

2) Construct histograms by quantizing the hue into different number of bins and voting with the corresponding Saturation value and vice versa.

3) Obtain DSCoP map from the value channel of HSV color space.

4) Form GLCM of DSCoP map by quantizing it into 16 levels.

5) Transform GLCM into a vector form.

6) Concatenate GLCM vector of step 5 with the histogram of step 2, and thus the final histogram is constructed as a feature vector.

**Part 2: Image retrieval using DSCoP + Modified Color Histogram**

**Input: Database query image**

**Output: Retrieved images after similarity measure**

1) Take the query image as input from the database.

2) Perform step-2 to step-6 in part 1 to extract the feature vector of the query image.

3) Using different similarity measures compute the similarity index of the query image vector with every database images.

4) Sort the similarity indices to produce the set of similar matching retrieved images as the final result.

### 3.2. Similarity Measure:

In content based image retrieval for retrieving and classifying images alongside color and texture feature computation, similarity measure is of same importance. The distance between the query image feature vector and feature of every image from the database in the feature space is given by the similarity measure which is performed after feature calculation. Indexing is then done based on this measure and sorting of the set of retrieved images is done based on the images with lower indices measures. Calculation of similarity matching is done using these five distance measures.

a. d1 distance:

$$\partial_{D,q_k} = \sum_{l=1}^{n} \left| \frac{\rho_d^k(l) - \rho_{q_k}(l)}{1 + \rho_d^k(l) + \rho_{q_k}(l)} \right| \quad \ldots\ldots\ldots (11)$$

b. Euclidean Distance

$$\partial_{D,q_k} = \left( \sum_{l=1}^{n} \left| (\rho_d^k(l) - \rho_{q_k}(l))^2 \right| \right)^{1/2} \quad \ldots\ldots\ldots (12)$$

c. Manhattan Distance

$$\partial_{D,q_k} = \sum_{l=1}^{n} \left| \rho_d^k(l) - \rho_{q_k}(l) \right| \quad \ldots\ldots\ldots (13)$$

d. Canberra Distance

$$\partial_{D,q_k} = \sum_{l=1}^{n} \left| \frac{\rho_d^k(l) - \rho_{q_k}(l)}{\rho_d^k(l) + \rho_{q_k}(l)} \right| \quad \ldots\ldots\ldots (14)$$

e. Chi-square Distance

$$\partial_{D,q_k} = \frac{1}{2} \sum_{l=1}^{n} \frac{(\rho_d^k(l) - \rho_{q_k}(l))^2}{\rho_d^k(l) + \rho_{q_k}(l)} \quad \ldots\ldots\ldots (15)$$

Here, the distance function for database D and query image $q_k$ is represented by $\partial_{D,q_k}$, n represents the length of the feature vector. The feature vector of $k^{th}$ database image and query image are $\rho_d^k(l)$ and $\rho_{q_k}(l)$ respectively.

## 4. Experimental Results and Analysis

In this paper, we have evaluated the performance of our method on 5 different datasets including texture image databases - MIT VisTex database and Salzburg texture database and natural scene databases Corel 1K, Corel 5K and Corel 10K. The superiority of the proposed method has been validated by evaluating the precision and recall rate and comparing it with existing methods on all these 5 datasets. Precision shows the relation between the total no. of relevant images retrieved for a given query image and the total no. of retrieved images from the database as in eqn. 16. Precision decreases as we gradually retrieve more images. The equation for determining the rate of precision may be given as:

$$\text{Precision Rate}(P_k, Q) = \frac{\text{Total no. of correct images retrieved from the database}}{\text{Total no. of images retrieved from the databases}(n)} \quad \ldots\ldots\ldots (16)$$

Here Q is the query image. $P_k$ represents the precision rate for category Q.

Another commonly used measure for determining accuracy is Recall. It can be defined as the probability of retrieving a correct relevant image by the query. For an image retrieval system, recall improves with increase in the number of images retrieved. It increases as more images are retrieved for different datasets. Recall can be viewed as the ratio of the total no. of relevant images retrieved for a given query image to the total no. of relevant images of that class from the database as in eqn. 17.

$$\text{Recall Rate}(R_k, Q) = \frac{\text{Total no of correct images retrieved from the database}}{\text{Total no of relevant images the databases}(N_k)} \quad \ldots \ldots \ldots (17)$$

here $N_k$ indicates number of images in each category of the database, i.e., the total number of relevant images in the database.

The average precision rate may be calculated as follows:

$$P_{avg}(M) = \frac{1}{j} \sum_{s=1}^{j} P_s \quad \ldots \ldots \ldots (18)$$

In eqn. 18 $P_{avg}(M)$ represents the average precision rate for category (M), where j is the total no of images in that category. Similarly, the recall rate for each category may be expressed as given in eqn. 19.

$$R_{avg}(M) = \frac{1}{j} \sum_{s=1}^{j} R_s \quad \ldots \ldots \ldots (19)$$

On similar terms, we can compute the total precision and total recall for our experiment using eqn. 20 and 21.

$$P_{total}(M) = \frac{1}{c} \sum_{i=1}^{C} P_{avg}(M) \quad \ldots \ldots \ldots (20)$$

$$R_{total}(M) = \frac{1}{c} \sum_{i=1}^{C} R_{avg}(M) \quad \ldots \ldots \ldots (21)$$

Here, C is the total no. of categories that is present in that particular database. Total Recall is also known as Average Recall Rate (ARR). The performance of the proposed method has been compared with a number of state-of-the-art methods. The list of abbreviations for these methods has been given in Table 1.

**Table 1: - Abbreviations of the different methods used:**

| Method | Name |
|---|---|
| Wavelet+colorhist | Discrete wavelet transform + RGB color histogram |
| CSLBP+colorhist | Center-symmetric local binary pattern + RGB color histogram |
| Joint LEP+colorhist | Joint histogram of color and LEP |
| Joint colorhist | Joint histogram of RGB color |
| LEPINV+colorhist | Local edge pattern for image retrieval + RGB color histogram |
| LEPSEG+colorhist | Local edge pattern for segmentation + RGB color histogram |
| LECoP+colorhist | Local extrema co-occurrence pattern + RGB color histogram |

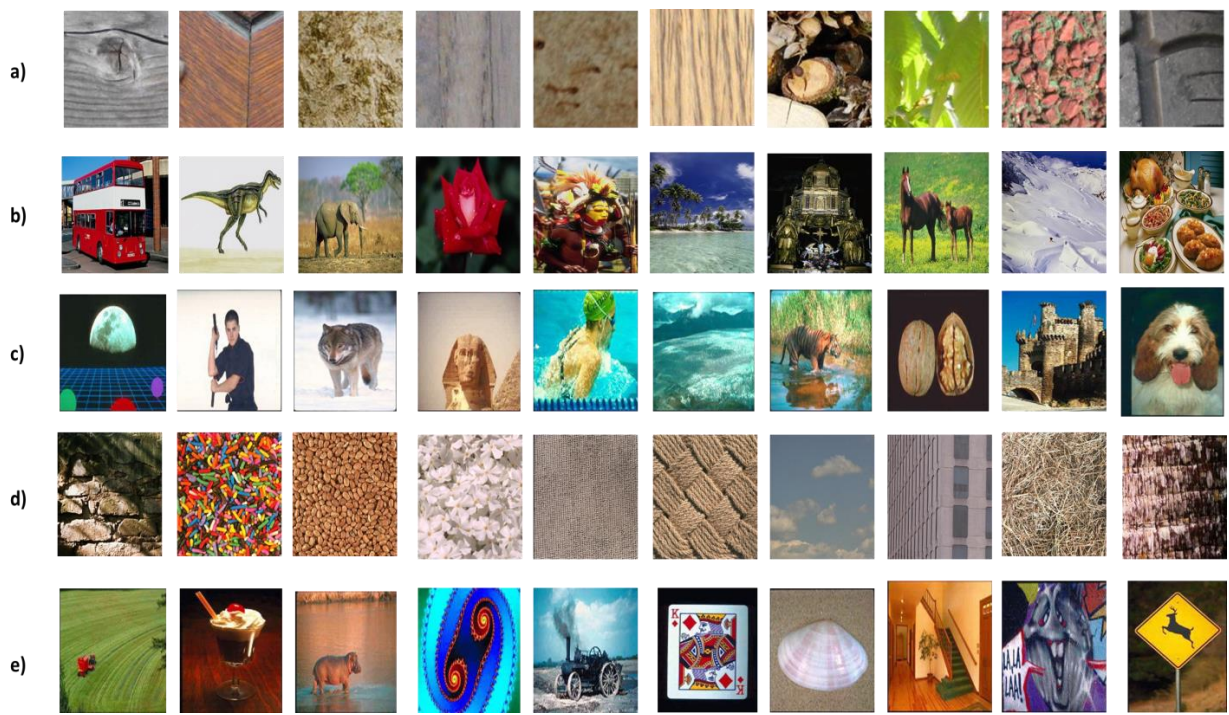

Fig. 5. Sample images from different datasets.

**Dataset 1:**

The first dataset used in our experiment is the Corel 1k database. It consists of 10 categories with 100 images in each category. Thus, there are a total of 1000 images in this database. The various categories of images in this database include Asians, buildings, beaches, elephant, flower, dinosaur, buses, mountains, hills and flood. Each image in this database has a size of 256×384 or a size of 384×256. Some sample images from this database are shown in Fig 5(b). The precision, recall and average retrieval rate have also been evaluated for this database. The precision and recall curves with different number of retrieved images for this dataset have been shown in Fig 6. For our experiment, we have initially retrieved 10 images and then increased the number of retrieved images in steps of 10 images at a time till the number of retrieved images becomes 100. In Fig. 7, the query image is represented by the first image of each row and the remaining images show the retrieved images for each query image. The comparative study of color and texture patterns for this dataset has been shown in Fig 8(a) and Fig 8(b) by studying the precision and recall considering each of the two individually.

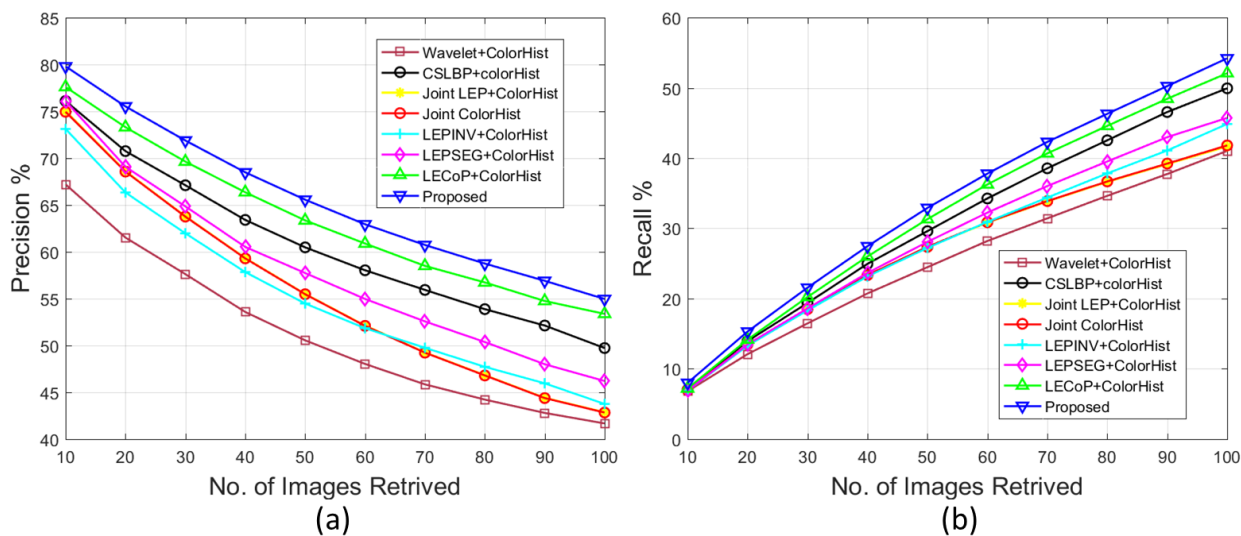

**Fig.6. Precision and recall curves with number of images retrieved for Corel-1K database**

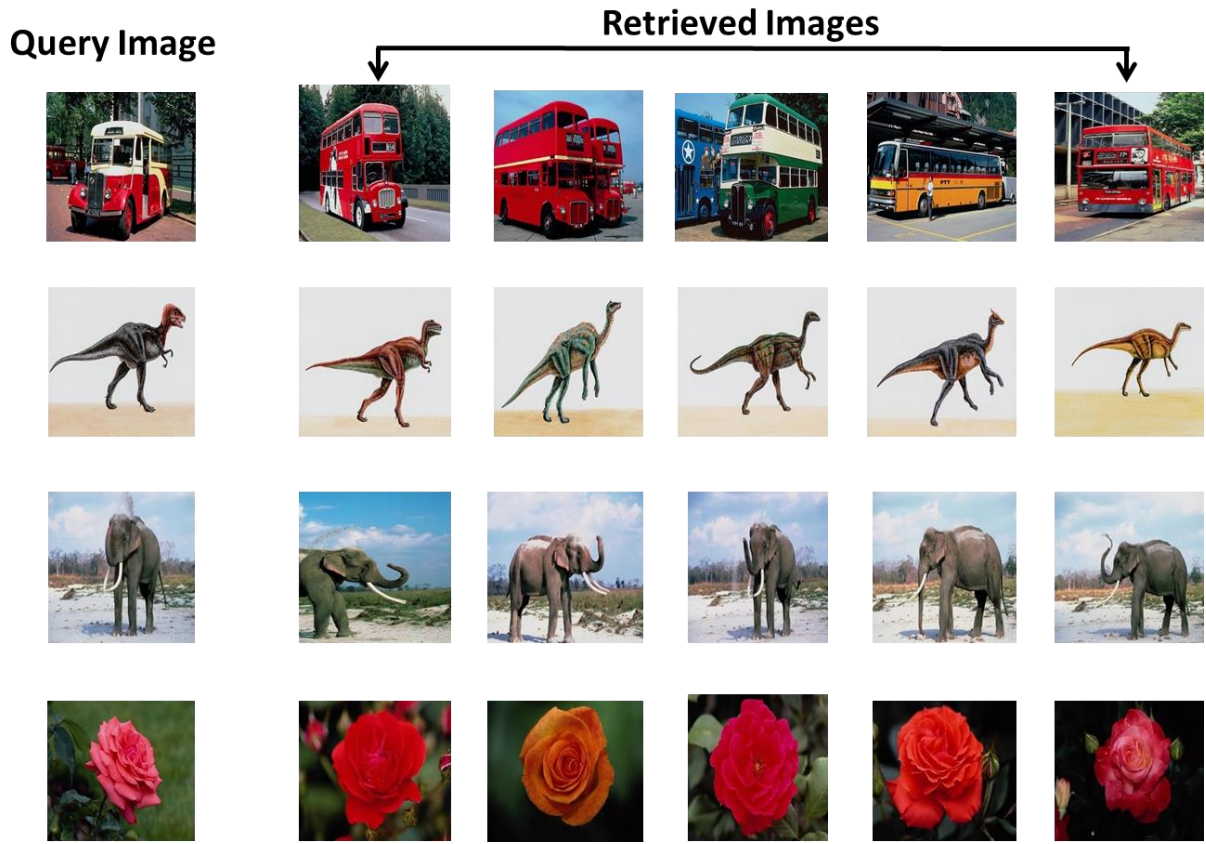

**Fig.7. Query image and retrieved images from Corel-1K dataset.**

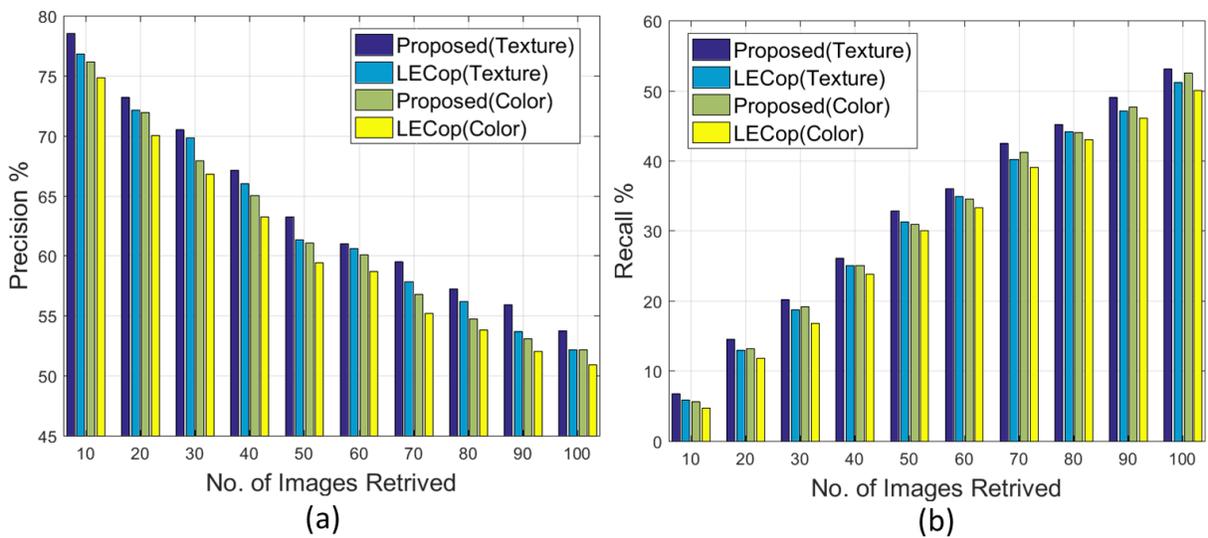

**Fig.8. Precision and recall values of proposed methods for Corel-1K database**

**Dataset 2:**

The second database that we have worked with in our experiment is the Corel 5K dataset. It consists of a total of 5000 images. There are a total of 50 categories and 100 images in each category. The dataset includes images of animals, e.g., bear, lion, fox, tiger, etc., human, buildings, paintings, natural scenes, fruits, cars, etc. In this experiment, we have retrieved 10 images initially. This has been increased till we retrieve 100 images to provide a fair comparison. The precision and recall curves for this dataset with varying number of images are shown in Fig. 9. Fig 5(e) shows some sample images for this dataset and the images retrieved corresponding to these sample images have been shown in Fig 10. The average retrieval rate for this dataset shown in Table 2 indicates that the proposed method performs better than state-of-the-art methods given in Table 1. We have shown in Fig. 11 that the texture pattern individually is more effective than the color pattern for CBIR.

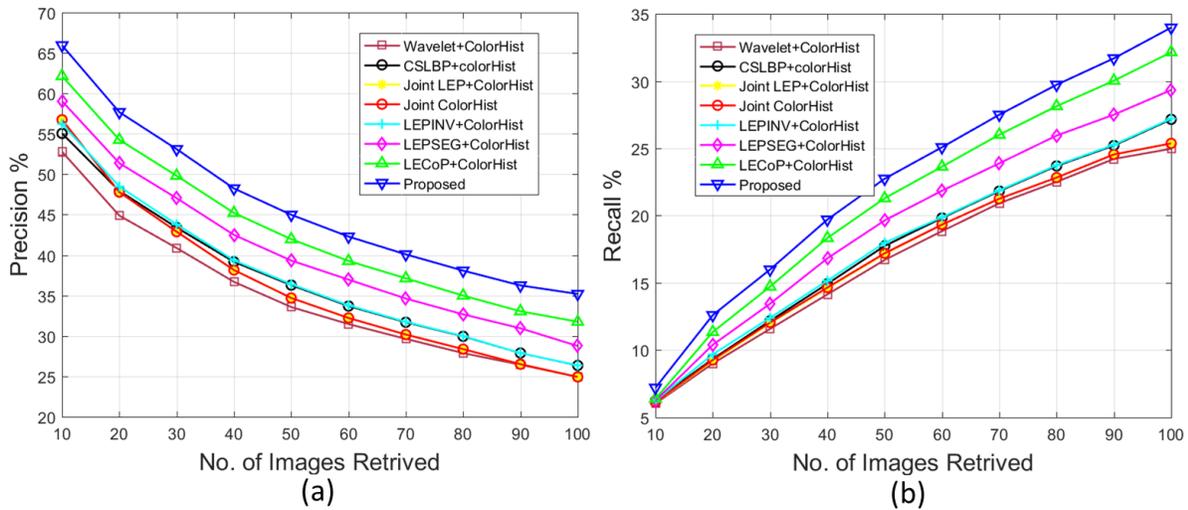

**Fig.9. Precision and recall curve with number of images retrieved for Corel-5K database**

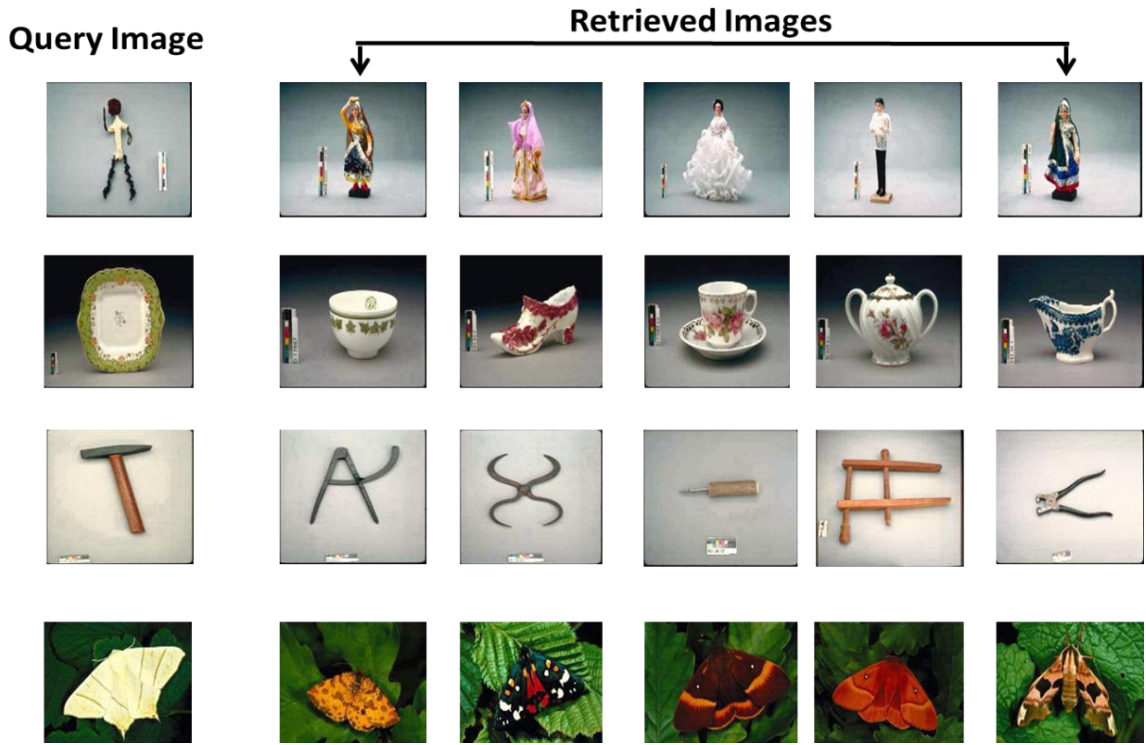

**Fig.10. Query image and retrieved images from Corel-5K dataset.**

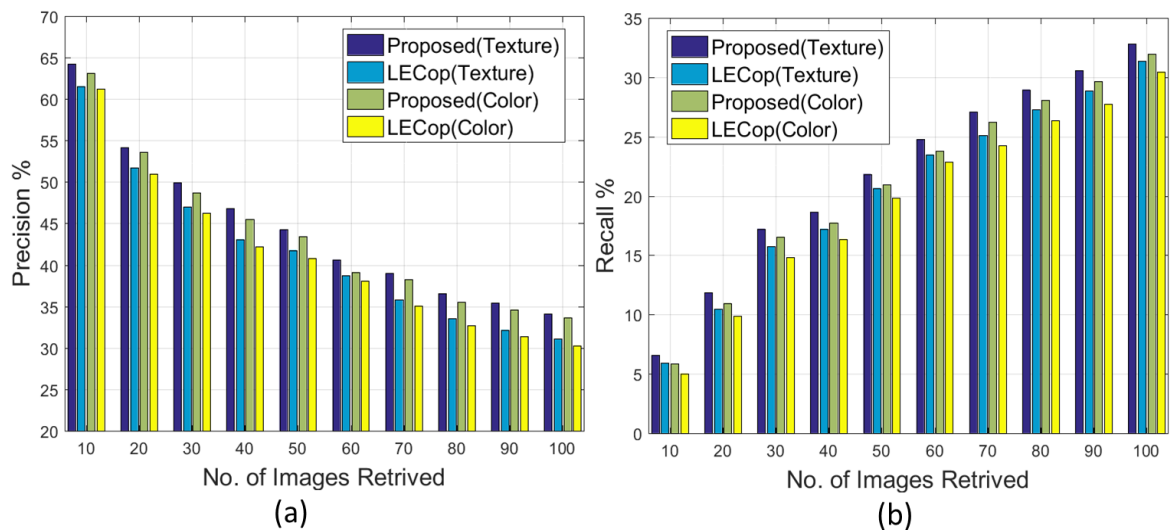

**Fig.11. Precision and recall value of proposed methods for Corel-5K database**

**Dataset 3:**

The third database we used in our experiment is the Corel 10K database[1]. It consists of 100 categories of images with 100 images in each category. This database is a continuation of Corel 5k database. It has images belonging to categories like buses, ships, texture, food, army, airplanes, furniture, oceans, cats, fishes, etc. Some sample images from this database are shown in Fig 5(c). The precision and recall curves for this dataset with different number of images retrieved for this dataset have been shown in Fig. 12. The Average Retrieval Rate for this dataset shows an improvement over existing methods given in Table 1. For experimental study, we have retrieved 10 images at first. We have then retrieved some more images from this dataset. Finally, we have retrieved 100 images from this dataset to provide a detailed comparison. In Fig. 13, the query image is represented by the first image of each row and the remaining images show the retrieved images for each query image. Fig.14. shows the comparative study of both texture and color feature of our method with LECoP.

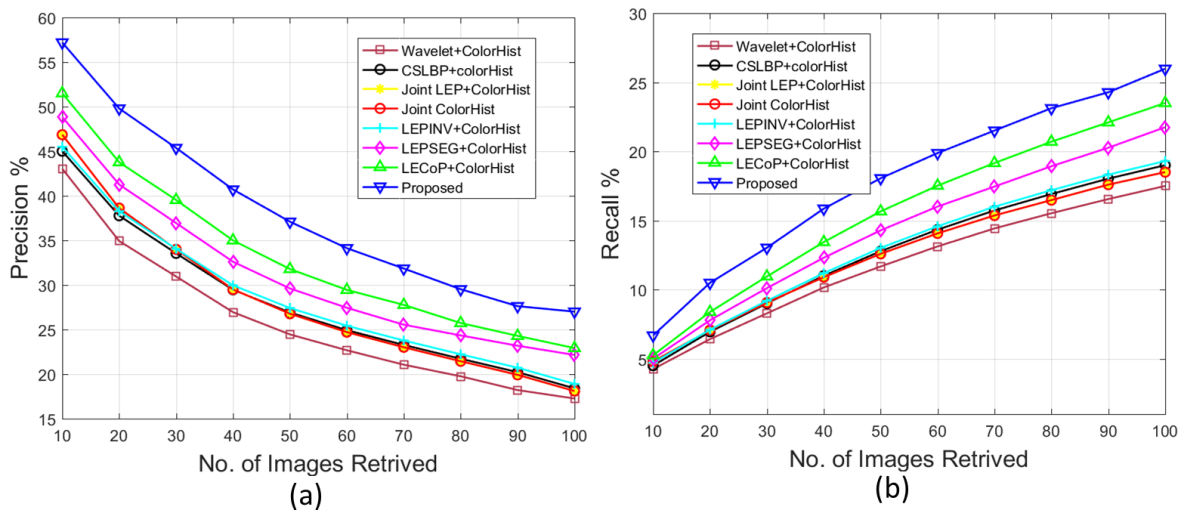

**Fig.12. Precision and recall curve with number of images retrieved for Corel-10K database**

---

[1] http://www.ci.gxnu.edu.cn/cbir/Dataset.aspx

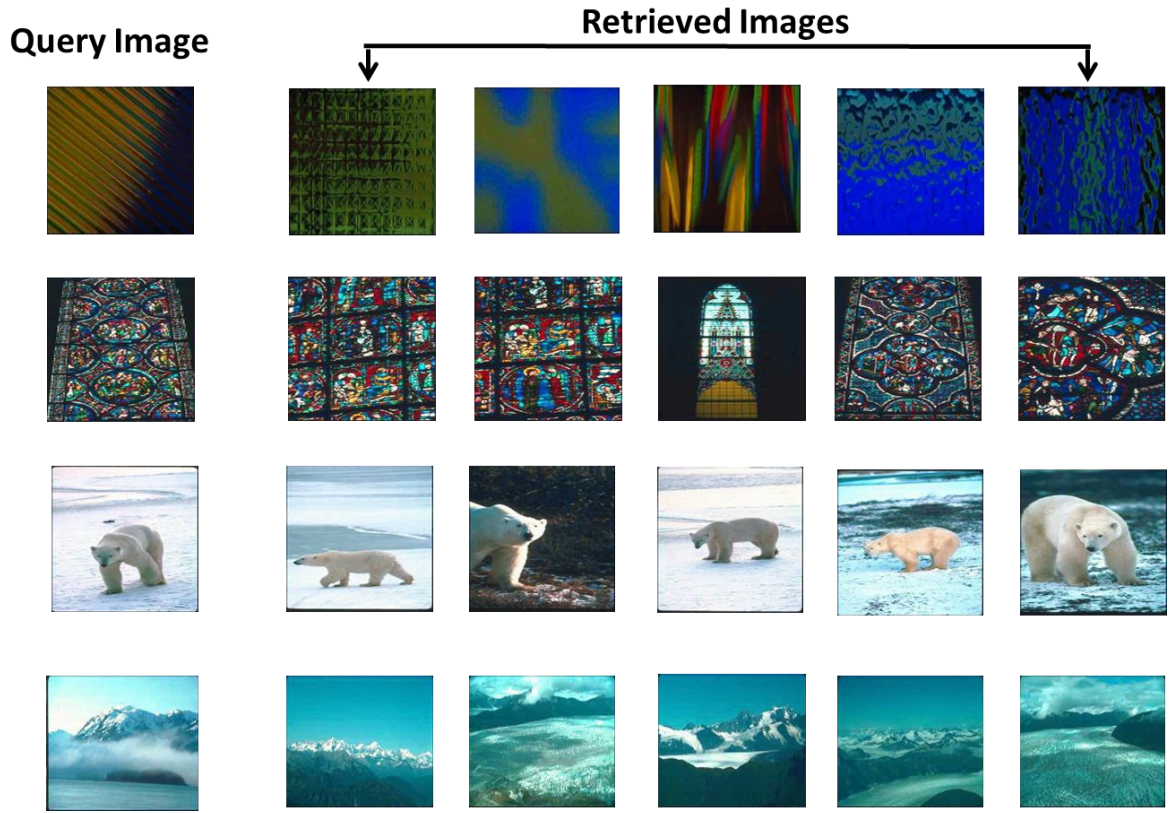

**Fig.13. Query image and retrieved images from Corel-10K dataset.**

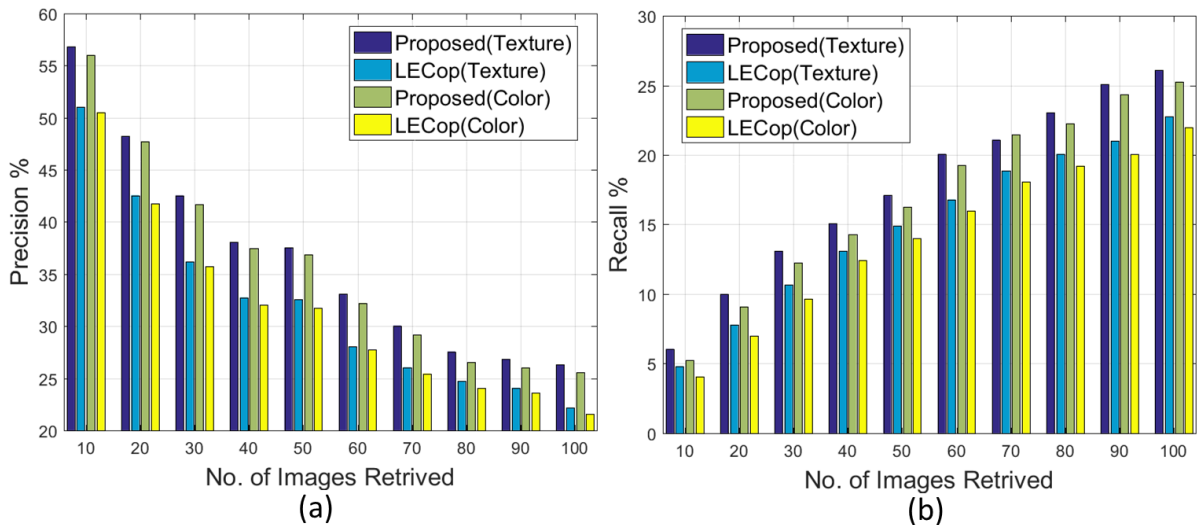

**Fig.14. Precision and recall value of proposed methods for Corel-10K database**

**Database 4:**

We have evaluated the performance of the proposed method on our fourth database the Salzburg texture (STex) database[2]. The database consists of 7616 images. This includes 476 categories with 16 images in each category. The sample images from this dataset are presented in Fig. 5(a). For STex dataset, we have retrieved 16 images to measure the precision and recall performance. To provide a detailed study, we have retrieved some more images which is reported with the help of precision and recall curves in Fig. 15(a) and Fig. 15(b). In Fig. 16, the query image is represented by the first images of each row and the remaining images show the retrieved images for each query image. Fig.17(a) and Fig.17(b). shows the comparative study of both texture and color feature of our method with LECoP.

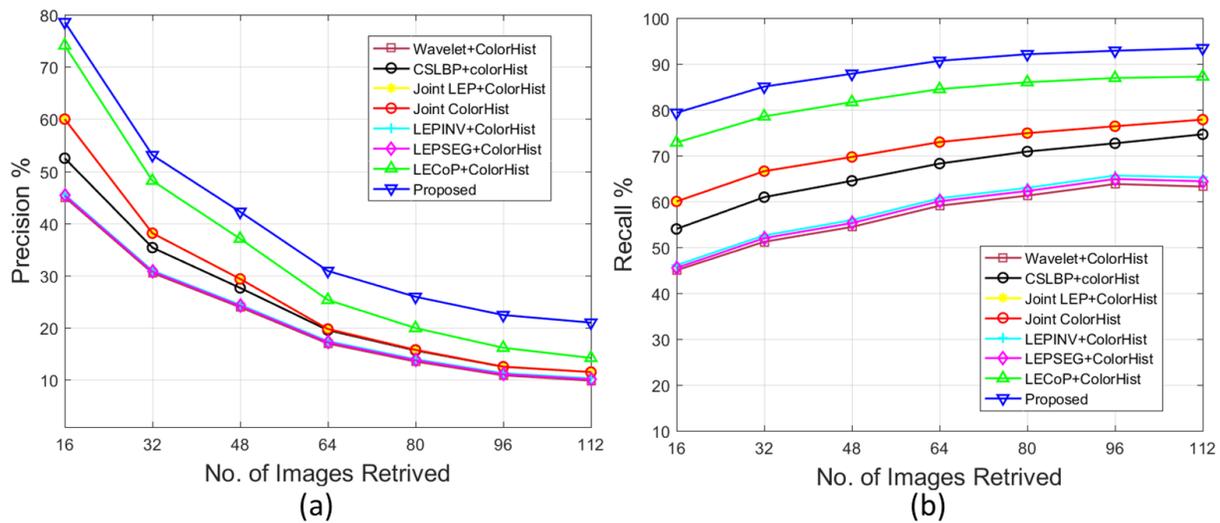

**Fig. 15. Precision and recall curve with number of images retrieved for STex database**

---

[2] http://www.wavelab.at/sources/STex/

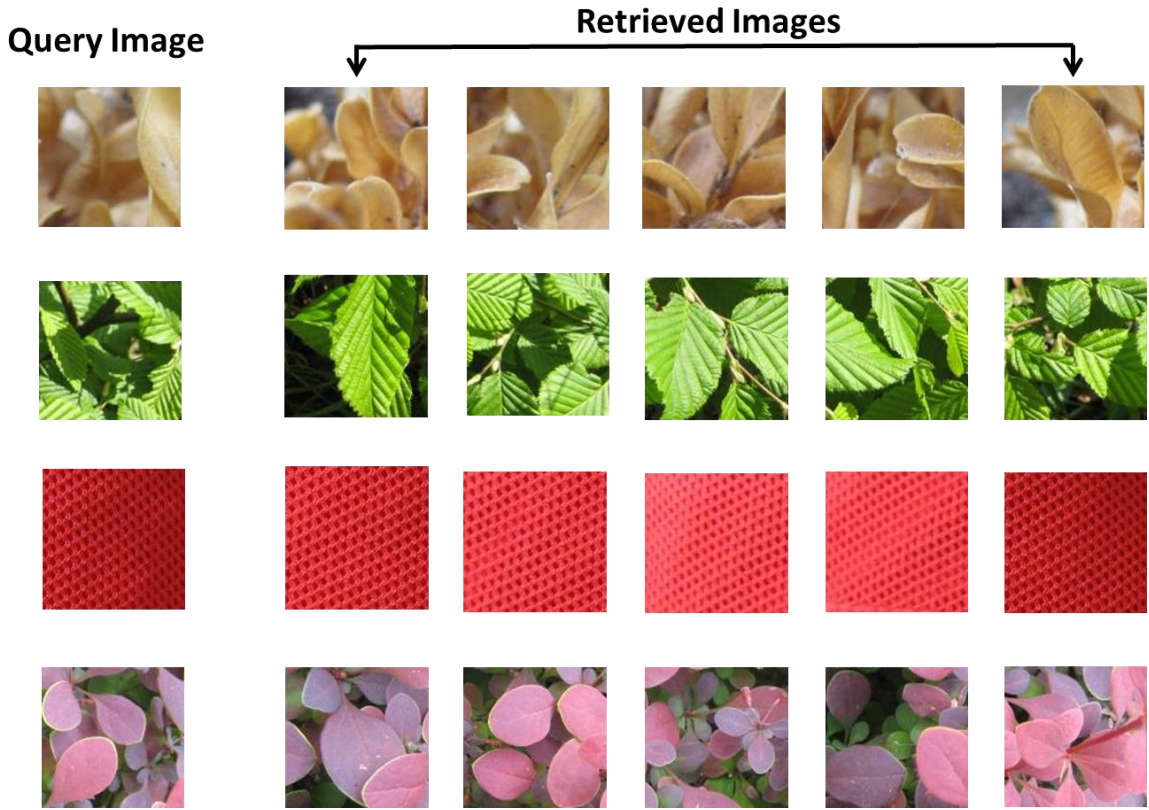

**Fig.16. Query image and retrieved images from STex database**

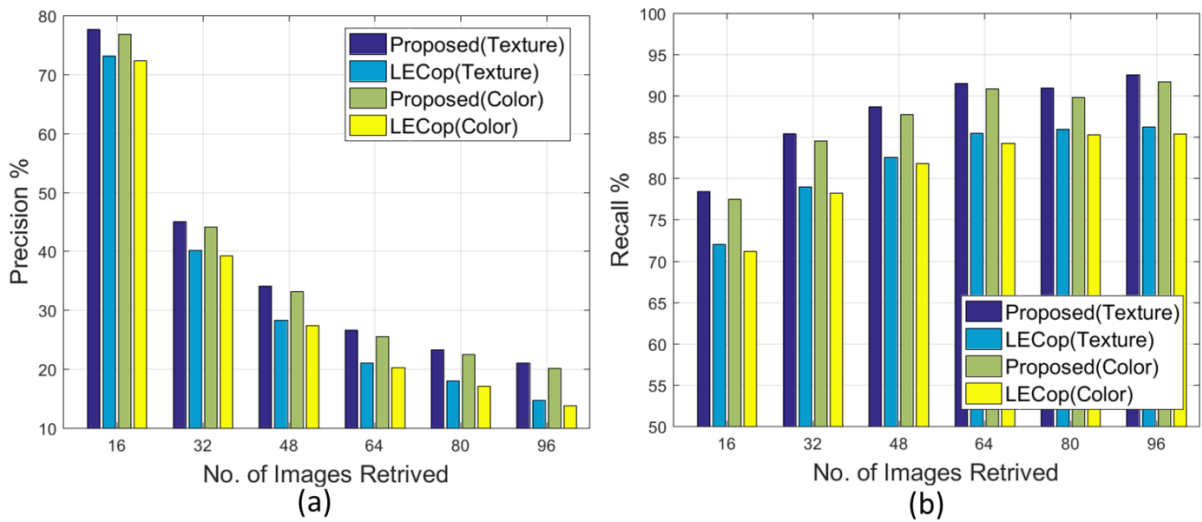

**Fig.17. Precision and recall value of proposed methods for STex database**

**Database 5:**

Finally, our method is tested on MIT-Vistex[3] database created by MIT Vision and Modeling Group. The dataset contains texture images of size 512×512. There are total 40 such gray-scale texture images. These images are again subdivided into images of size 128× 128. So, the dataset is divided into 40 different types of images with 16 images of each type. In this experiment, initially 16 images are retrieved and then the number of images retrieved is increased by 16. The maximum number of retrieved images is 96. The Precision Rate and Recall Rate for all images in the database are calculated and compared with the methods in Table 1. A graph in support of our observations is shown in Fig. 18. Some sample images from our dataset is shown in Fig. 5(d) and some query images and their corresponding retrieved images are shown in Fig. 19. Our proposed texture and color feature performs better than the texture and color feature of LECoP as shown in Fig. 20(a) and Fig. 20(b).

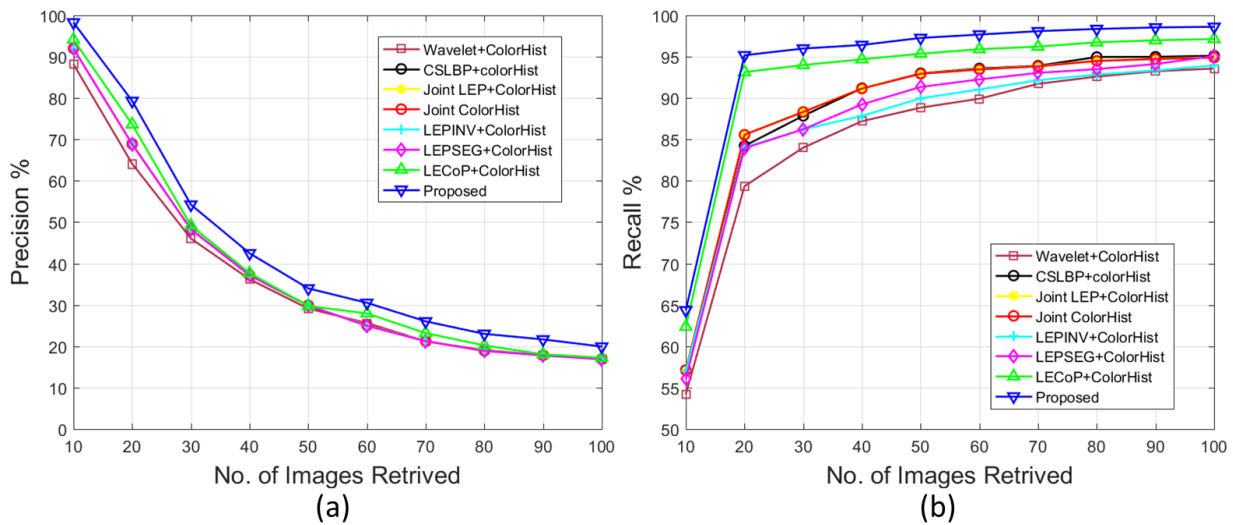

**Fig.18. Precision and recall curve with number of images retrieved for MIT-Vistex database**

---

[3] MIT Vision and Modeling Group, Cambridge, Vision texture, available online: http://vismod.media.mit.edu/pub/.

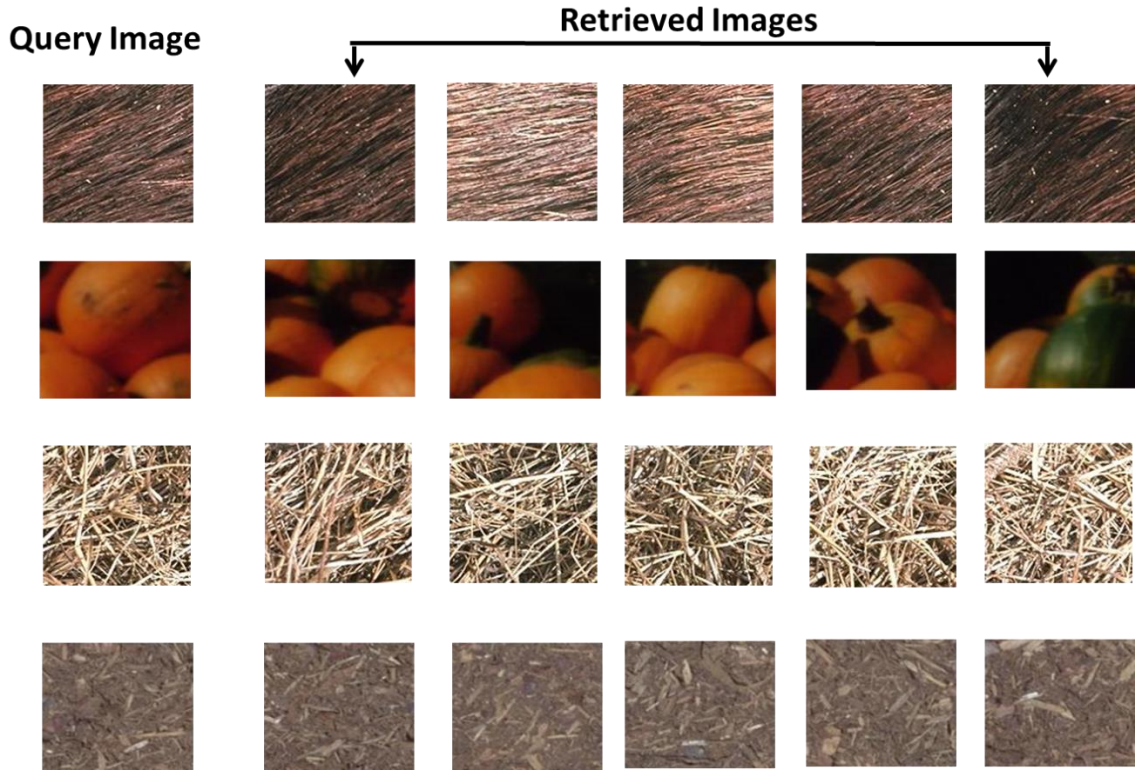

**Fig.19. Query image and retrieved images from MIT-Vistex dataset.**

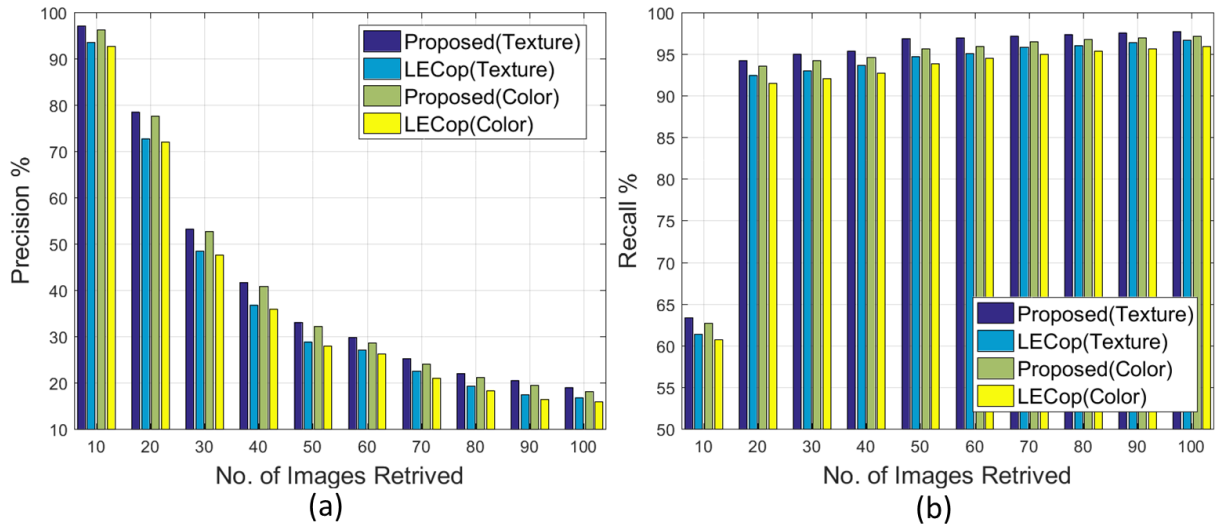

**Fig.20. Precision and recall value of proposed methods for MIT-Vistex database**

The average retrieval rate for this method has also been determined as in Table 2. It has been compared with recent methods for different datasets. We have also studied the image retrieval and feature extraction time of several state-of-the-art feature descriptors and compared the same with our method. The feature vector length and image retrieval time of the recently developed techniques have been shown in Table 3. Various similarity metrics have been considered for performance evaluation as in eqn.11-15. We have shown the performance of the proposed method using different similarity or distance metrics as given in Table 4. The performance varies with different distance metrics on different datasets. However, the best performance on all datasets is obtained for d1 distance metric. We have also shown a comparative performance with LECoP (Local Extrema Co-occurrence pattern) by studying the texture and the color patterns separately. Both the patterns show an improvement over the corresponding patterns of LECoP. However, the texture pattern turns out to be more effective than the texture pattern for all datasets as indicated by the precision recall values shown with the help of bar graphs. In Table 5, the individual importance of hue, saturation and value in HSV color space is analyzed with different values of quantization levels of hue and saturation components for all databases.

**Table 2: Average Retrieval Rate for STex and MIT-Vistex datasets**

| Different Feature | STex | MIT-Vistex |
|---|---|---|
| Wavelet+colorhist | 44.21 | 74.94 |
| CSLBP+colorhist | 52.48 | 80.12 |
| Joint LEP+colorhist | 58.84 | 81.41 |
| Joint colorhist | 58.62 | 81.54 |
| LEPINV+colorhist | 47.23 | 79.65 |
| LEPSEG+colorhist | 45.56 | 78.84 |
| LECoP+colorhist | 74.15 | 92.54 |
| Proposed | 76.38 | 93.67 |

**Table 3: Feature retrieval time, extraction time and feature length using different methods**

| Different Feature | Feature Length | Feature Extraction Time | Retrieval Time |
|---|---|---|---|
| Wavelet+colorhist | 192+24=216 | 0.0746 | 0.48 |
| CSLBP+colorhist | 16+24=40 | 0.1205 | 0.47 |
| Joint LEP+colorhist | $16 \times 8 \times 8 \times 8$=8192 | 0.1656 | 2.48 |
| Joint colorhist | $8 \times 8 \times 8$=512 | 0.0349 | 0.54 |
| LEPINV+colorhist | 72+24=96 | 0.0694 | 0.48 |
| LEPSEG+colorhist | 512+24=536 | 0.0232 | 0.55 |
| LECoP($H_{18}S_{10}V_{256}$) | 18+10+256=284 | 0.2391 | 0.52 |
| LECoP($H_{18}S_{20}V_{256}$) | 18+20+256=294 | 0.2406 | 0.52 |

| LECoP($H_{36}S_{10}V_{256}$) | 36+10+256=302 | 0.2415 | 0.52 |
|---|---|---|---|
| LECoP($H_{36}S_{20}V_{256}$) | 36+20+256=312 | 0.2428 | 0.54 |
| LECoP($H_{72}S_{10}V_{256}$) | 72+10+256=338 | 0.2435 | 0.56 |
| LECoP($H_{72}S_{20}V_{256}$) | 72+20+256=348 | 0.2447 | 0.56 |
| PM($H_{18}S_{10}V_{256}$) | 18+10+256=284 | 0.2378 | 0.51 |
| PM ($H_{18}S_{20}V_{256}$) | 18+20+256=294 | 0.2384 | 0.52 |
| PM($H_{36}S_{10}V_{256}$) | 36+10+256=302 | 0.2393 | 0.52 |
| PM($H_{36}S_{20}V_{256}$) | 36+20+256=312 | 0.2402 | 0.54 |
| PM($H_{72}S_{10}V_{256}$) | 72+10+256=338 | 0.2415 | 0.55 |
| PM($H_{72}S_{20}V_{256}$) | 72+20+256=348 | 0.2428 | 0.55 |

**Table 4: Comparative study with different distance matrics**

| Distance measure | Corel 1K | Corel 5K | Corel 10K | STex | MIT-Vistex |
|---|---|---|---|---|---|
| d1 | 51.35 | 31.06 | 23.14 | 90.18 | 99.54 |
| Euclidean | 40.56 | 19.05 | 14.54 | 75.61 | 92.79 |
| Manhattan | 47.65 | 23.75 | 18.40 | 84.74 | 97.15 |
| Canberra | 45.88 | 27.15 | 21.46 | 86.84 | 98.87 |
| Chi Square | 48.25 | 24.15 | 19.57 | 83.25 | 96.58 |

**Table 5: Precision and Recall values of our method with different quantization schemes for all databases**

|  | Corel 1k | | Corel 5k | | Corel 10k | | STex | | MIT-Vistex | |
|---|---|---|---|---|---|---|---|---|---|---|
|  | APR | ARR | APR | ARR | APR | ARR | APR | ARR | APR | ARR |
| HSV(18,10,256) | 79.52 | 51.78 | 64.16 | 32.36 | 53.70 | 24.49 | 73.83 | 89.99 | 93.74 | 99.13 |
| HSV(18,20,256) | 78.42 | 54.55 | 64.30 | 31.81 | 53.67 | 24.13 | 74.45 | 90.63 | 94.19 | 99.32 |
| HSV(36,10,256) | 79.70 | 51.90 | 62.76 | 31.47 | 52.38 | 23.80 | 24.57 | 90.64 | 93.34 | 99.18 |
| HSV(36,20,256) | 79.86 | 52.92 | 64.09 | 32.04 | 53.72 | 24.25 | 75.35 | 91.23 | 94.15 | 99.33 |
| HSV(72,10,256) | 79.22 | 51.97 | 62.43 | 30.73 | 52.42 | 23.33 | 74.52 | 90.56 | 92.72 | 99.18 |
| HSV(72,20,256) | 79.78 | 52.97 | 61.66 | 29.92 | 52.06 | 23.41 | 75.21 | 91.10 | 93.38 | 99.36 |

## 5. Conclusion

This paper presents a novel approach towards content based image retrieval by proposing a novel descriptor by combining the color and texture information. The texture descriptor is named Diagonally Symmetric Local Binary Co-occurrence Pattern since it effectively captures the co-occurrence relationship between the symmetric neighbor pairs about the left and right diagonals of an image. The color descriptor focuses on capturing the inter-channel relationship between the H and S channels of the HSV color space by quantizing the H channel into bins and voting with Saturation value and replicating the process for the S channel. The texture descriptor developed in this paper effectively captures the co-occurrence relationship between the neighbor pairs symmetric about the principle and counter diagonal of an image. The method has been evaluated on texture image databases - MIT VisTex database and Salzburg texture database and natural scene databases Corel 1K, Corel 5K and Corel 10K. The result obtained has been compared with existing techniques by calculating the precision and recall values for all of them. The proposed method turns out to be better than the existing approaches in terms of both precision and recall. The feature vector length and image retrieval rate are also competitive with most approaches. Thus, in real time systems this image retrieval technique is quite effective and efficient.

# References

[1] S. R. Dubey, S. K. Singh, and R. K. Singh, "Local diagonal extrema pattern: A new and efficient feature descriptor for CT image retrieval," *IEEE Signal Process. Lett.*, vol. 22, no. 9, pp. 1215–1219, 2015.
[2] R. M. Haralick and K. Shanmugam, "Textural Features for Image Classification," *IEEE Trans. Syst. Man. Cybern.*, vol. 3, no. 6, pp. 610–621, 1973.
[3] J. Zhang, G. L. Li, and S. W. He, "Texture-based image retrieval by edge detection matching GLCM," in *Proceedings - 10th IEEE International Conference on High Performance Computing and Communications, HPCC 2008*, 2008, pp. 782–786.
[4] M. Partio, B. Cramariuc, M. Gabbouj, and A. Visa, "Rock texture retrieval using gray level co-occurrence matrix," *Proc. 5th Nord. Signal …*, 2002.
[5] F. Roberti de Siqueira, W. Robson Schwartz, and H. Pedrini, "Multi-scale gray level co-occurrence matrices for texture description," *Neurocomputing*, vol. 120, pp. 336–345, 2013.
[6] Y. Li, C. Zhou, B. Geng, C. Xu, and H. Liu, "A comprehensive study on learning to rank for content-based image retrieval," *Signal Processing*, vol. 93, no. 6, pp. 1426–1434, 2013.
[7] S. Fadaei, R. Amirfattahi, and M. R. Ahmadzadeh, "Local derivative radial patterns: A new texture descriptor for content-based image retrieval," *Signal Processing*, vol. 137, pp. 274–286, 2017.
[8] W. Li, H. Pan, P. Li, X. Xie, and Z. Zhang, "A medical image retrieval method based on texture block coding tree," *Signal Process. Image Commun.*, vol. 59, pp. 131–139, 2017.
[9] A. K. Tiwari, V. Kanhangad, and R. B. Pachori, "Histogram refinement for texture descriptor based image retrieval," *Signal Process. Image Commun.*, vol. 53, pp. 73–85, 2017.
[10] P. Banerjee, A. K. Bhunia, A. Bhattacharyya, P. P. Roy, and S. Murala, "Local Neighborhood Intensity Pattern: A new texture feature descriptor for image retrieval," *arXiv Prepr. arXiv1709.02463*, 2017.
[11] C. Palm, "Color texture classification by integrative Co-occurrence matrices," *Pattern Recognit.*, vol. 37, no. 5, pp. 965–976, 2004.
[12] S. Jeong, C. S. Won, and R. M. Gray, "Image retrieval using color histograms generated by Gauss mixture vector quantization," *Comput. Vis. Image Underst.*, vol. 94, no. 1–3, pp.

44–66, 2004.

[13] G. Pass, R. Zabih, and J. Miller, "Comparing images using color coherence vectors," *Proc. fourth ACM Int. Conf. Multimed. (MULTIMEDIA '96)*, pp. 1–14, 1998.

[14] M. Subrahmanyam, Q. M. Jonathan Wu, R. P. Maheshwari, and R. Balasubramanian, "Modified color motif co-occurrence matrix for image indexing and retrieval," *Comput. Electr. Eng.*, vol. 39, no. 3, pp. 762–774, 2013.

[15] A. Baraldi and F. Parmiggiani, "An investigation of the textural characteristics associated with gray level cooccurrence matrix statistical parameters," *IEEE Trans. Geosci. Remote Sens.*, vol. 33, no. 2, pp. 293–304, 1995.

[16] V. Kovalev and M. Petrou, "Multidimensional Co-occurrence Matrices for Object Recognition and Matching," *Graph. Model. IMAGE Process.*, vol. 58, no. 3, pp. 187–197, 1996.

[17] L. S. Davis, S. a Johns, and J. K. Aggarwal, "Texture analysis using generalized co-occurrence matrices.," *IEEE Trans. Pattern Anal. Mach. Intell.*, vol. 1, no. 3, pp. 251–259, 1979.

[18] A. Vadivel, S. Sural, and A. K. Majumdar, "An integrated color and intensity co-occurrence matrix," *Pattern Recognit. Lett.*, vol. 28, no. 8, pp. 974–983, 2007.

[19] J. Huang, S. R. Kumar, M. Mitra, W.-J. Zhu, and R. Zabih, "Image indexing using color correlograms," in *Computer Vision and Pattern Recognition, 1997. Proceedings., 1997 IEEE Computer Society Conference on*, 1997, pp. 762–768.

[20] J. Huang, S. R. Kumar, and M. Mitra, "Combining supervised learning with color correlograms for content-based image retrieval," *Proc. fifth ACM Int. Conf. Multimed. - Multimed. '97*, pp. 325–334, 1997.

[21] S. T. Park, K. Seo, and D. Jang, "Expert system based on artificial neural networks for content-based image retrieval," *Expert Syst. Appl.*, vol. 29, no. 3, pp. 589–597, 2005.

[22] N. Jhanwar, S. Chaudhuri, G. Seetharaman, and B. Zavidovique, "Content based image retrieval using motif cooccurrence matrix," *Image Vis. Comput.*, vol. 22, no. 14, pp. 1211–1220, 2004.

[23] S. K. Vipparthi and S. K. Nagar, "Multi-joint histogram based modelling for image indexing and retrieval," *Comput. Electr. Eng.*, vol. 40, no. 8, pp. 163–173, 2014.

[24] L. Balmelli and A. Mojsilovic, "Wavelet domain features for texture description, classification and replicability analysis," in *Image Processing, 1999. ICIP 99. Proceedings. 1999 International Conference on*, 1999, vol. 4, pp. 440–444.

[25] S. Ardizzoni, I. Bartolini, and M. Patella, "Windsurf: region-based image retrieval using wavelets," *Proc. Tenth Int. Work. Database Expert Syst. Appl. DEXA 99*, pp. 167–173, 1999.

[26] J. Z. Wang, G. Wiederhold, O. Firschein, and S. X. Wei, "Content-based image indexing and searching using Daubechies' wavelets," *Int. J. Digit. Libr.*, vol. 1, no. 4, pp. 311–328, 1997.

[27] H. A. Moghaddam, T. T. Khajoie, A. H. Rouhi, and M. S. Tarzjan, "Wavelet correlogram: A new approach for image indexing and retrieval," *Pattern Recognit.*, vol. 38, no. 12, pp. 2506–2518, 2005.

[28] B. S. Manjunath, "Texture features for browsing and retrieval of image data," *IEEE Trans. Pattern Anal. Mach. Intell.*, vol. 18, no. 8, pp. 837–842, 1996.

[29] a. Ahmadian and a. Mostafa, "An efficient texture classification algorithm using Gabor wavelet," *Proc. 25th Annu. Int. Conf. IEEE Eng. Med. Biol. Soc. (IEEE Cat. No.03CH37439)*, vol. 1, no. OCTOBER 2003, pp. 930–933, 2003.

[30] H. Abrishami Moghaddam and M. Nikzad Dehaji, "Enhanced Gabor wavelet correlogram feature for image indexing and retrieval," *Pattern Anal. Appl.*, vol. 16, no. 2, pp. 163–177, 2013.

[31] T. Ojala, M. Pietikainen, and T. Maenpaa, "Multiresolution gray-scale and rotation invariant texture classification with local binary patterns," *IEEE Trans. Pattern Anal. Mach. Intell.*, vol. 24, no. 7, pp. 971–987, 2002.

[32] Z. Guo, L. Zhang, and D. Zhang, "A completed modeling of local binary pattern operator for texture classification.," *IEEE Trans. Image Process.*, vol. 19, no. 6, pp. 1657–1663, 2010.

[33] V. Takala, T. Ahonen, and M. Pietikainen, "Block-based methods for image retrieval using Local Binary Patterns," in *Lecture Notes in Computer Science*, 2005, vol. 3540, pp. 882–891.

[34] S. Liao, M. W. K. Law, and A. C. S. Chung, "Dominant local binary patterns for texture classification," *IEEE Trans. Image Process.*, vol. 18, no. 5, pp. 1107–1118, 2009.

[35] M. Heikkilä, M. Pietikäinen, and C. Schmid, "Description of interest regions with center-symmetric local binary patterns," *Comput. Vision, Graph. Image Process.*, vol. 2, pp. 58–69, 2006.

[36] Y. He, N. Sang, and C. Gao, "Multi-structure local binary patterns for texture classification," *Pattern Anal. Appl.*, pp. 1–13, 2012.

[37] X. Qian, X. S. Hua, P. Chen, and L. Ke, "PLBP: An effective local binary patterns texture descriptor with pyramid representation," *Pattern Recognit.*, vol. 44, no. 10–11, pp. 2502–2515, 2011.

[38] L. Tlig, M. Sayadi, and F. Fnaiech, "A new fuzzy segmentation approach based on S-FCM type 2 using LBP-GCO features," *Signal Process. Image Commun.*, vol. 27, no. 6, pp. 694–708, 2012.

[39] G. A. Papakostas, D. E. Koulouriotis, E. G. Karakasis, and V. D. Tourassis, "Moment-based local binary patterns: A novel descriptor for invariant pattern recognition applications," *Neurocomputing*, vol. 99, pp. 358–371, 2013.

[40] S. Murala, R. P. Maheshwari, and R. Balasubramanian, "Directional local extrema patterns: a new descriptor for content based image retrieval," *Int. J. Multimed. Inf. Retr.*, vol. 1, no. 3, pp. 191–203, 2012.

[41] S. R. Dubey, S. K. Singh, and R. K. Singh, "Local Bit-Plane Decoded Pattern: A Novel Feature Descriptor for Biomedical Image Retrieval," *IEEE J. Biomed. Heal. Informatics*, vol. 20, no. 4, pp. 1139–1147, 2016.

[42] C. H. Yao and S. Y. Chen, "Retrieval of translated, rotated and scaled color textures," *Pattern Recognit.*, vol. 36, no. 4, pp. 913–929, 2002.

[43] S. Murala and Q. M. J. Wu, "Local mesh patterns versus local binary patterns: Biomedical image indexing and retrieval," *IEEE J. Biomed. Heal. Informatics*, vol. 18, no. 3, pp. 929–938, 2014.

[44] I. Hamouchene and S. Aouat, "A New Texture Analysis Approach for Iris Recognition," *AASRI Procedia*, vol. 9, pp. 2–7, 2014.

[45] X. Tan and B. Triggs, "Enhanced local texture feature sets for face recognition under difficult lighting conditions," *IEEE Trans. Image Process.*, vol. 19, no. 6, pp. 1635–1650, 2010.

[46] X. Wu, J. Sun, G. Fan, and Z. Wang, "Improved local ternary patterns for automatic target recognition in infrared imagery," *Sensors (Switzerland)*, vol. 15, no. 3, pp. 6399–6418, 2015.

[47] J. Ren, X. Jiang, and J. Yuan, "Noise-resistant local binary pattern with an embedded error-correction mechanism," *IEEE Trans. Image Process.*, vol. 22, no. 10, pp. 4049–4060, 2013.

[48] Y. Zhao, W. Jia, R. X. Hu, and H. Min, "Completed robust local binary pattern for texture classification," *Neurocomputing*, vol. 106, pp. 68–76, 2013.

[49] S. Murala, R. P. Maheshwari, and R. Balasubramanian, "Local tetra patterns: A new feature descriptor for content-based image retrieval," *IEEE Trans. Image Process.*, vol. 21, no. 5, pp. 2874–2886, 2012.

[50] I. Jeena Jacob, K. G. Srinivasagan, and K. Jayapriya, "Local Oppugnant Color Texture Pattern for image retrieval system," *Pattern Recognit. Lett.*, vol. 42, no. 1, pp. 72–78, 2014.

[51] S. Murala and Q. M. Jonathan Wu, "Spherical symmetric 3D local ternary patterns for natural, texture and biomedical image indexing and retrieval," *Neurocomputing*, vol. 149, no. PC, pp. 1502–1514, 2015.

[52] M. Verma, B. Raman, and S. Murala, "Local extrema co-occurrence pattern for color and texture image retrieval," *Neurocomputing*, vol. 165, pp. 255–269, 2015.

[53] G.-H. Liu and J.-Y. Yang, "Content-based image retrieval using color difference histogram," *Pattern Recognit.*, vol. 46, no. 1, pp. 188–198, 2013.

[54] E. Walia and A. Pal, "Fusion framework for effective color image retrieval," *J. Vis. Commun. Image Represent.*, vol. 25, no. 6, pp. 1335–1348, 2014.

[55] Z. Lu, X. Jiang, and A. Kot, "A novel LBP-based Color descriptor for face recognition," in *Acoustics, Speech and Signal Processing (ICASSP), 2017 IEEE International Conference on*, 2017, pp. 1857–1861.

[56] A. Ahmadian, A. Mostafa, M. Abolhassani, and Y. Salimpour, "A texture classification method for diffused liver diseases using Gabor wavelets.," *Conf. Proc. IEEE Eng. Med. Biol. Soc.*, vol. 2, no. c, pp. 1567–1570, 2005.

[57] L. Nanni, A. Lumini, and S. Brahnam, "Local binary patterns variants as texture descriptors for medical image analysis," *Artif. Intell. Med.*, vol. 49, no. 2, pp. 117–125, 2010.

[58] J. Ning, L. Zhang, D. Zhang, and C. Wu, "Robust Object Tracking Using Joint Color-Texture Histogram," *Int. J. Pattern Recognit. Artif. Intell.*, vol. 23, no. 7, pp. 1245–1263, 2009.

[59] S. Moore and R. Bowden, "Local binary patterns for multi-view facial expression recognition," *Comput. Vis. Image Underst.*, vol. 115, no. 4, pp. 541–558, 2011.